\documentclass{article}

\usepackage[preprint]{arxiv}

\makeatletter
\renewcommand{\@toptitlebar}{}
\renewcommand{\@bottomtitlebar}{}
\makeatother

\usepackage[utf8]{inputenc} 
\usepackage[T1]{fontenc}    
\usepackage{hyperref}       
\usepackage{url}            
\usepackage{booktabs}       
\usepackage{amsfonts}       
\usepackage{nicefrac}       
\usepackage{microtype}      
\usepackage{xcolor}         
\usepackage{amsmath}
\usepackage{tabularx}
\usepackage{graphicx}     
\usepackage{array}  
\usepackage{colortbl}
\usepackage{wrapfig} 
\usepackage{fontawesome5} 
\usepackage{subcaption}  

\usepackage{multirow}
\usepackage{amsmath}
\usepackage{mathtools}              
\usepackage{enumitem}   
\usepackage{graphicx}               
\usepackage{tabularx}               
\usepackage{booktabs}
\usepackage{amsmath}
\usepackage{stmaryrd}
\usepackage{xcolor}
\usepackage{soul}

\usepackage{graphicx}               
\usepackage{tabularx}               
\usepackage{booktabs}
\usepackage{amsmath}
\usepackage{stmaryrd}
\usepackage{xcolor}
\usepackage{soul}

\newcolumntype{C}{>{\centering\arraybackslash}X}
\usepackage{multirow}               
\usepackage{diagbox}                
\usepackage{hhline}                 
\usepackage{color}                  
\usepackage{amsmath}                
\usepackage{mathtools}              
\usepackage{enumitem} 
\usepackage{booktabs}
\usepackage{extarrows}
\usepackage{makecell}
\usepackage{xparse}
\usepackage{soul}
\usepackage{xcolor}
\usepackage[linesnumbered,ruled,vlined]{algorithm2e}
\usepackage{adjustbox} 

\usepackage{dsfont}

\newcommand{\mat}[1]{\ensuremath{\mathbf{\uppercase{#1}}}} 

\definecolor{RoseQuartzBg}{HTML}{F7CAC9}
\definecolor{ForestGreen}{HTML}{228b22}
\definecolor{RoseQuartz}{HTML}{F5A798}
\definecolor{Serenity}{HTML}{92A8D1}
\definecolor{OrangeRed}{rgb}{1.0, 0.27, 0.0}
\definecolor{Red}{rgb}{1.0, 0.0, 0.0}
\definecolor{forestgreen}{rgb}{0.13, 0.55, 0.13}
\definecolor{Turquoise}{HTML}{0F4C81}
\definecolor{columbiablue}{rgb}{0.61, 0.87, 1.0}
\definecolor{Gray}{gray}{0.9}

\usepackage{xparse}
\usepackage{footmisc}

\NewDocumentCommand{\zhiyang}{ mO{} }{\textcolor{ForestGreen}{\textsuperscript{\textit{zhiyang}}\textsf{\textbf{\small[#1]}}}}

\newcommand{\modelname}{BLIP3o-NEXT} 
\newcommand{\dataname}{{BLIP3o-60k}}

\usepackage[most]{tcolorbox}

\tcbset{
  finding style/.style={
    enhanced,
    colback=gray!10,           
    colframe=gray!60,         
    boxrule=0.8pt,             
    arc=2mm,                   
    left=1mm, right=1mm,       
    top=1mm, bottom=1mm,
    before skip=10pt, after skip=10pt,
    fontupper=\normalfont,     
    fonttitle=\normalfont\bfseries, 
  }
}

\newtcolorbox{findingbox}[1][]{finding style,#1}

\title{\modelname{}: Next Frontier of Native Image Generation}

\author{
\textbf{Jiuhai Chen}\textsuperscript{1,2*} \quad
\textbf{Le Xue}\textsuperscript{1*}  \quad
\textbf{Zhiyang Xu}\textsuperscript{3*} \quad
\textbf{Xichen Pan}\textsuperscript{4} \quad
\textbf{Shusheng Yang}\textsuperscript{4} \\
\textbf{Can Qin}\textsuperscript{1}\quad \textbf{An Yan}\textsuperscript{1}\quad \textbf{Honglu Zhou}\textsuperscript{1}\quad \textbf{Zeyuan Chen}\textsuperscript{1}\quad  \textbf{Lifu Huang}\textsuperscript{6}\quad 
\textbf{Tianyi Zhou}\textsuperscript{2}\\
\textbf{Junnan Li}\textsuperscript{1} \quad
\textbf{Silvio Savarese}\textsuperscript{1\ddag} \quad
\textbf{Caiming Xiong}\textsuperscript{1\ddag}  \quad
\textbf{Ran Xu}\textsuperscript{1\ddag} \\
\\
\textsuperscript{1}Salesforce Research \\
\textsuperscript{2}University of Maryland \quad
\textsuperscript{3}Virginia Tech \quad
\textsuperscript{4}New York University \quad
\textsuperscript{6} UC Davis \\
\\
\textsuperscript{*}Equal Contribution. \quad
\textsuperscript{\ddag}Corresponding Authors.
}

\begin{document}

\maketitle

\begin{abstract}
{

We present \modelname{}, a fully open-source foundation model in the BLIP3 series that advances the next frontier of native image generation. \modelname{} unifies text-to-image generation and image editing within a single architecture, demonstrating strong image generation and image editing capabilities. 
In developing the state-of-the-art native image generation model, we identify four key insights: (1) Most architectural choices yield comparable performance; an architecture can be deemed effective provided it scales efficiently and supports fast inference;
(2) The successful application of reinforcement learning can further push the frontier of native image generation;
(3) Image editing still remains a challenging task, yet instruction following and the consistency between generated and reference images can be significantly enhanced through post-training and data engine; 
(4) Data quality and scale continue to be decisive factors that determine the upper bound of model performance.
Building upon these insights, \modelname{} leverages an Autoregressive + Diffusion architecture in which an autoregressive model first generates discrete image tokens conditioned on multimodal inputs, whose hidden states are then used as conditioning signals for a diffusion model to generate high-fidelity images. This architecture integrates the reasoning strength and instruction following of autoregressive models with the fine-detail rendering ability of diffusion models, achieving a new level of coherence and realism. 
Extensive evaluations of various text-to-image and image-editing benchmarks show that \modelname{} achieves superior performance over existing models.
}

\end{abstract}


\begin{center}
    \renewcommand{\arraystretch}{1.2}
    \begin{tabular}{rll}
        \faGlobe & \textbf{Website} & \url{https://jiuhaichen.github.io/BLIP3o-NEXT.github.io}\\
        \faGithub & \textbf{Code} & \url{https://github.com/JiuhaiChen/BLIP3o}\\
    \end{tabular}
\end{center}

\newpage


\section{Introduction}
Recently image generation has become a cornerstone of modern artificial intelligence, empowering models to not only synthesize photorealistic images from textual descriptions but also edit existing images with remarkable precision and semantic consistency~\cite{gpt4o,pan2025transfer,chen2025blip3,wu2025omnigen2,deng2025emerging,wu2025qwen}. These capabilities have redefined how machines interpret, represent, and communicate visual information, enabling a wide range of creative and practical applications, from content creation and design to scientific visualization and simulation.

In this paper, we introduce \modelname{}, a novel image generation foundation model that advances the frontier of multimodal learning through innovations in model architecture, multi-task pretraining, reinforcement learning, and comprehensive data engineering. \modelname{} adopts a hybrid Autoregressive + Diffusion architecture~\cite{pan2025transfer,chen2025blip3}. The autoregressive model~\cite{bai2025qwen2} takes a user prompt together with a set of reference images (for image editing) and generates a sequence of discrete image tokens conditioned on the multimodal input. The diffusion model~\cite{xie2024sana} subsequently leverages the final hidden states of generated discrete tokens as conditioning signals to synthesize the final image. This architecture combines the semantic compositionality and global structure understanding captured by the autoregressive model with the fine-grained detail rendering capability of diffusion models, enabling \modelname{} to produce images that are both semantically coherent and photorealistically detailed. 


\begin{figure}[!t]
\centering
\includegraphics[width=\linewidth]{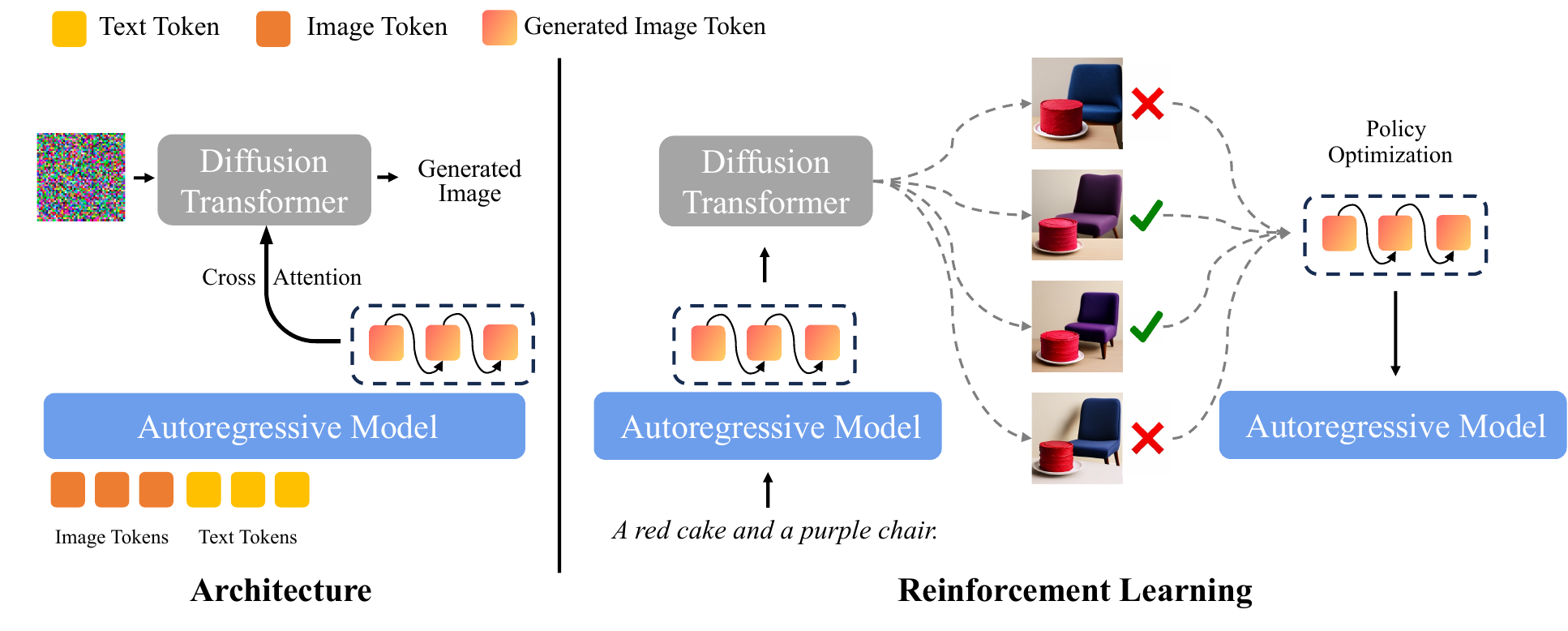}
\caption{
The architecture of \modelname{} (left) and its reinforcement learning pipeline (right). \modelname{} adopts an Autoregressive (AR) + Diffusion design, where the AR module autoregressively generates image conditions for the diffusion model. The model is jointly optimized with both AR and diffusion objectives. During reinforcement learning, rollouts are rendered from the diffusion transformer, and policy optimization is performed directly on the AR model, enabling seamless integration with existing RL infrastructures originally developed for language models.
}
\label{fig:architecture}
\end{figure}


 The autoregressive model is trained on three primary tasks: (1) text-to-image generation, (2) input image reconstruction, and (3) image editing, which together equip the model with the fundamental capabilities required for diverse downstream tasks. In the post-training stage, beyond fine-tuning on carefully curated high-quality datasets~\cite{chen2025blip3,wu2025omnigen2,chen2023sharegpt4vimprovinglargemultimodal}, we perform extensive reinforcement learning (RL) to further enhance the model’s text rendering quality and instruction following ability. We propose an efficient RL framework specifically tailored for the \modelname{} architecture by leveraging the discrete image tokens. This design allows seamless integration with existing RL infrastructures originally developed for language models, while achieving performance comparable to recent flow-based RL approaches.

Taking advantage of the Autoregressive + Diffusion design, \modelname{} exhibits great flexibility in integrating the VAE features~\cite{dcae} of reference images for editing tasks. Through empirical experiments, we find that concatenating VAE features to (1) hidden states generated by the autoregressive model as conditions, and (2) the random noise as initial input in the diffusion process, achieves the highest consistency and visual fidelity. Additionally, powered by the strong multimodal reasoning and understanding capability of its autoregressive backbone, \modelname{} can faithfully follow user instructions for image editing.


In summary, \modelname{} makes the following key contributions:

\begin{itemize}
\item \textbf{A novel and scalable Autoregressive + Diffusion architecture} that advances the next frontier of native image generation.
\item \textbf{An efficient reinforcement learning method for image generation} that can be seamlessly integrated with existing RL infrastructures for language models, improving  text rendering and instruction following abilities.
\item \textbf{Systematic studies on improving consistency in image editing}, including strategies for integrating VAE features from reference images.

\item \textbf{Strong performance across diverse benchmarks}, comprehensive evaluation on text-to-image generation benchmarks and image-editing benchmarks reveals that \modelname{} consistently outperform existing models.
\end{itemize}

\textit{To support future research and uphold the open-source philosophy of the BLIP3 family, we fully release \modelname{}, including pretrained and post-trained model weights, datasets, detailed training and inference code, and evaluation pipelines, ensuring complete reproducibility. We hope that our work will foster continued progress in native image generation.}

\section{Overview of Architectures for Native Image Generation}


\subsection{Autoregressive + Diffusion}

In recent developments in native image generation, approaches that integrate autoregressive models with diffusion models continue to achieve state-of-the-art performance. In these frameworks, the autoregressive backbone encodes the input prompt, whether text or other modalities, and produces conditions for the diffusion model to generate the final image. Early explorations include Emu2~\cite{sun2024generative} and Seed-X~\cite{ge2024seed}. Following the success of GPT-4o~\cite{gpt4o}, subsequent works such as MetaQuery~\cite{pan2025transfer}, UniGen~\cite{unigen}, and BLIP3-o~\cite{chen2025blip3} push the direction further, while recent models like Qwen-Image~\cite{wu2025qwen}, Gemini Nano Banana and Seedream~\cite{seedream2025seedream} continue to push the performance frontier. The advantage of Autoregressive + Diffusion is its \textbf{simplicity} and \textbf{scalability}, and also leveraging the strengths of existing autoregressive models, such as transferring instruction following and reasoning capabilities into image generation procedure.

Within this framework, the question is how to derive the conditioning signal from the autoregressive model for the diffusion model. Existing approaches can be broadly grouped into the following categories: (1) The autoregressive model directly generates continuous embeddings (e.g., CLIP representations), which are then used as conditions for the diffusion model, similar to Emu2~\cite{sun2024generative} and MetaMorph~\cite{tong2024metamorph}. (2) The autoregressive model compresses all input information, including text and images, into a fixed number of learnable query, as demonstrated by Seed-X~\cite{ge2024seed}, MetaQuery~\cite{pan2025transfer}, and BLIP3-o~\cite{chen2025blip3}. (3) The autoregressive model encodes both image and text inputs, and the hidden states from the autoregressive model are directly used as the conditioning signal for the diffusion model, as in OmniGen2~\cite{wu2025omnigen2}, Qwen-Image~\cite{wu2025qwen}, and MANZANO~\cite{li2025manzano}. 

Since continuous embeddings and learnable query tokens generate conditioning signals by sampling from AR models, they tend to perform better on image generation tasks requiring reasoning capability compared with directly using hidden states~\cite{tang2025unilip}. However, both continuous embeddings and learnable query compress all conditioning information into a fixed number of tokens, which inevitably limits their representational capacity. For example, Emu2~\cite{sun2024generative} and BLIP3-o~\cite{chen2025blip3} usually use 64 tokens. In contrast, using hidden states offers a more flexible and efficient way for encoding the condition. 


\subsection{Mixture of Transformers}

LMFusion~\cite{shi2024llamafusion} and BAGEL~\cite{deng2025emerging} adopt a different architecture for native image generation. At a high level, they employ two transformer experts to separately process understanding and generation information, while tokens from different modalities interact through shared multimodal self-attention within each transformer block. 
Although this structure facilitates richer information sharing across modalities, it faces challenges in flexibility and scalability, and suffers from high inference latency.

\subsection{\modelname{} Architecture}


In \modelname{}, we still adopt an Autoregressive + Diffusion architecture, where an autoregressive model first predicts discrete image tokens from multimodal inputs, and the hidden states of these tokens are subsequently used to condition a diffusion model for high-fidelity image synthesis. Unlike previous approaches, where the autoregressive model takes in and generates continuous image embeddings~\cite{ge2024seed,tong2024metamorph,pan2025transfer,chen2025blip3}, our autoregressive model operates on discrete visual tokens. Specifically, each image is first encoded using the SigLIP2 model~\cite{han2025vision}, and the resulting continuous embeddings are quantized into a finite vocabulary of tokens, yielding 729 discrete tokens per image given the image resized to 384x384. Conditioned on a text prompt (or reference images for image editing), the autoregressive model is trained through next-token prediction over discrete image tokens. The diffusion model is then applied on top of the autoregressive outputs to refine fine-grained details. Specifically, the hidden states of the predicted tokens serve as conditioning signals for the diffusion model to diffuse the final VAE features, which produce the final high-fidelity images, as illustrated in Figure \ref{fig:architecture} (left).

During training, \modelname{} is optimized with two objectives:
\begin{equation}
\mathcal{L} = \mathcal{L}_{\text{CE}} + \lambda \, \mathcal{L}_{\text{diff}},
\end{equation}
where $\mathcal{L}_{\text{CE}}$ denotes the cross-entropy loss over text and discrete image tokens in autoregressive model, $\mathcal{L}_{\text{diff}}$ is the diffusion loss, and $\lambda$ is a balancing weight. Due to computational constraints, we initialize the autoregressive model with Qwen3~\cite{yang2025qwen3} and the diffusion transformer with SANA1.5~\cite{xie2024sana}, resulting in a total of approximately 3B parameters. For training, we employ BLIP3o-Pretrain~\cite{chen2025blip3} as the pretraining corpus and BLIP3o-60K~\cite{chen2025blip3} as the instruction tuning dataset.

\subsection{Discussion}

From GPT-4o, where an autoregressive model produces discrete image tokens and is refined by a diffusion model, to more recent models such as Qwen-Image~\cite{wu2025qwen}, architectures are increasingly dominated by diffusion backbones. For example, Qwen-Image integrates a 7B vision language model with a 20B diffusion transformer. In practice, GPT-4o exhibits higher inference latency, whereas advances in accelerating diffusion transformers~\cite{shao2025rayflow, ren2024hyper, lin2025diffusion} have enabled diffusion-centric models to achieve both greater scalability and faster inference efficiency.

\begin{findingbox}
 Under Autoregressive + Diffusion framework, we observe that most architectures deliver comparable performance, with minor design variations yielding marginal differences. Therefore, an architecture can be considered effective, as long as it remains simple, scalable, and supports fast inference.
\end{findingbox}


\section{Image Generation with Reinforcement Learning}

Reinforcement learning (RL) has been extensively explored in large language models, but remains relatively underexplored in the image generation domain. GPT-4o~\cite{gpt4o} is the representative native image generation model to successfully integrate RL. In this work, we primarily discuss two directions for applying RL to image generation: (1) applying RL to the autoregressive mode under the \modelname{} framework, and (2) applying RL to the diffusion model, for example Flow-GRPO\cite{liu2025flow}.

\subsection{RL for Autoregressive Model}


Applying RL to the autoregressive model is a natural extension of language modeling. The key advantage of this approach is its compatibility with existing RL frameworks developed for language models, allowing most training and inference infrastructures to be easily adapted for image generation training.

We use the Group Relative Policy Optimization (GRPO) algorithm~\cite{shao2024deepseekmath} under \modelname{} framework. For each text prompt $p \sim \mathcal{D}$, the previous policy $\pi_{\theta_{old}}$, i.e., the autoregressive model, samples a group of $G$ trajectories $\{o_1, \ldots, o_G\}$. Each of the trajectory consists of 729 discrete image tokens for an image.
These trajectories are then decoded by a frozen diffusion model to generate the corresponding images $\{I_1, I_2, \ldots, I_G\}$, each assigned a reward score $\{r_1, \ldots, r_G\}$ by the reward function. The rewards are normalized across the group to compute the advantages $A_i$, following the GRPO procedure. During training, the diffusion model remains frozen, while the policy model $\pi_\theta$ is optimized by maximizing the following objective:  
\begin{align}\label{eq:grpo}
& \mathcal{L}_{GRPO}(\theta) =
\mathbb{E}_{p \sim \mathcal{D}, \, \{o_i\}_{i=1}^G \sim \pi_{\theta_{old}}(\cdot|p)}  \\
& \quad \frac{1}{G} \sum_{i=1}^G \Bigg(
\min \!\left(
\frac{\pi_\theta(o_i|p)}{\pi_{\theta_{old}}(o_i|p)} A_i,\;
\text{clip}\!\left( \frac{\pi_\theta(o_i|p)}{\pi_{\theta_{old}}(o_i|p)}, 1-\epsilon, 1+\epsilon \right) A_i
\right) \nonumber  \quad - \beta \, \mathbb{D}_{KL}\!\left(\pi_\theta \,\|\, \pi_{\theta_{ref}}\right)
\Bigg).
\end{align}

Here, $\epsilon$ and $\beta$ are hyperparameters, $\pi_{\theta_{ref}}$ is a fixed reference policy, and $\mathbb{D}_{KL}\!\left(\pi_\theta \,\|\, \pi_{\theta_{ref}}\right)$  is a KL divergence penalty that constrains policy updates relative to the reference policy. RL training can be seen in Figure~\ref{fig:architecture} right.

\begin{table*}[t]
\centering
\small
\setlength{\tabcolsep}{1pt}
\begin{tabular}{lccccccl}
\toprule
\textbf{Model} & \textbf{Single Obj.} & \textbf{Two Obj.} & \textbf{Counting} & \textbf{Colors} & \textbf{Position} & \textbf{Color Attri.} & \textbf{Overall} \\
\midrule
FLUX.1-dev (12B)~\cite{labs2025flux} & 0.98 & 0.93 & 0.75 & 0.93 & 0.68 & 0.65 & 0.82 \\
OmniGen2 (7B)~\cite{wu2025omnigen2} &1.00& 0.95& 0.64 &0.88& 0.55& 0.76& 0.80 \\
Qwen-Image (27B)~\cite{wu2025qwen} &
0.99 & 0.92 & 0.89& 0.88& 0.76& 0.77& 0.87 \\
Metaqueries XL (7B)~\cite{pan2025transfer} & -- & -- & -- & -- & -- & -- & 0.80$^{\dagger}$  \\
BAGEL (14B)~\cite{deng2025emerging} & 0.98 & 0.95 & 0.84 & 0.95 & 0.78 & 0.77 & 0.88$^{\dagger}$ \\
BLIP3o (8B)~\cite{chen2025blip3}  & -- & -- & -- & -- & -- & -- & 0.84 \\
\rowcolor{green!10} BLIP3o-NEXT-GRPO-GenEval (3B) & 0.99 & 0.95 & 0.88 & 0.90 & 0.92 & 0.79 & 0.91 \\
\bottomrule
\end{tabular}
\caption{Quantitative results on GenEval benchmarks. ($^{\dagger}$ denotes the rewritten prompts.)}
\label{tab:geneval}
\end{table*}


\subsection{RL for Diffusion Model}

Instead of autoregressive model, RL can also be applied to the diffusion model. Qwen-Image~\cite{wu2025qwen} is the representative model to successfully apply both DPO~\cite{rafailov2023direct} and Flow-GRPO~\cite{liu2025flow} within a native image generation model.

After the DPO stage, Qwen-Image uses additional fine-grained reinforcement learning with GRPO, following the Flow-GRPO framework~\cite{liu2025flow}.  
Conditioned on the hidden state of the text prompt $p$ from the autoregressive module, the flow model generates a set of $G$ candidate images $\{x_0^i\}_{i=1}^G$ along with their trajectories $\{x_T^i, x_{T-1}^i, \dots, x_0^i\}_{i=1}^G$.  
The training objective of Flow-GRPO is consistent with Equation~\ref{eq:grpo}.  

A key distinction from RL in autoregressive models is trajectory sampling.  
In Flow-GRPO, trajectories $\{x_T^i, \dots, x_0^i\}_{i=1}^G \sim \pi_\theta$ are sampled according to the flow matching dynamics:
\begin{equation}
dx_t = v_t \, dt,
\end{equation}
where $v_t = v_\theta(x_t, t, p)$ is the velocity predicted by the model.  
However, this deterministic formulation lacks stochasticity and thus fails to support adequate exploration.  
To address this, Flow-GRPO reformulates trajectory sampling as a stochastic differential equation (SDE), injecting randomness into the process:  
\begin{equation}
dx_t = \left( v_t + \frac{\sigma_t^2}{2t}(x_t + (1-t)v_t) \right) dt + \sigma_t dw_t,
\end{equation}

\noindent where $dw_t$ denotes Brownian motion and $\sigma_t$ controls the magnitude of randomness.

In addition to Flow-GRPO, DanceGRPO~\cite{xue2025dancegrpo} extends GRPO to a broader range of visual generation tasks, including text-to-image, text-to-video, and image-to-video. It reformulates both diffusion sampling and rectified flows within the framework of stochastic differential equations, enabling GRPO to generalize across different architectures and training paradigms. MixGRPO~\cite{li2025mixgrpo} further improves computational efficiency by combining SDE and ODE based formulations.

\subsection{Reward Models}

Reward functions can be broadly divided into two categories.
(1) Verifiable rewards, such as GenEval~\cite{ghosh2023geneval} for the composition of multiple objects and OCR-based evaluation for visual text rendering.
(2) Model-based rewards, including PickScore~\cite{kirstain2023pick}, ClipScore~\cite{hessel2021clipscore}, HPSv2.1~\cite{wu2023human}, ImageReward~\cite{xu2023imagereward}, and UnifiedReward~\cite{wang2025unified}, which assess image quality, image–text alignment, and human preference.

\subsection{Experiments}
\begin{figure}[!t]
\centering
\includegraphics[width=\linewidth]{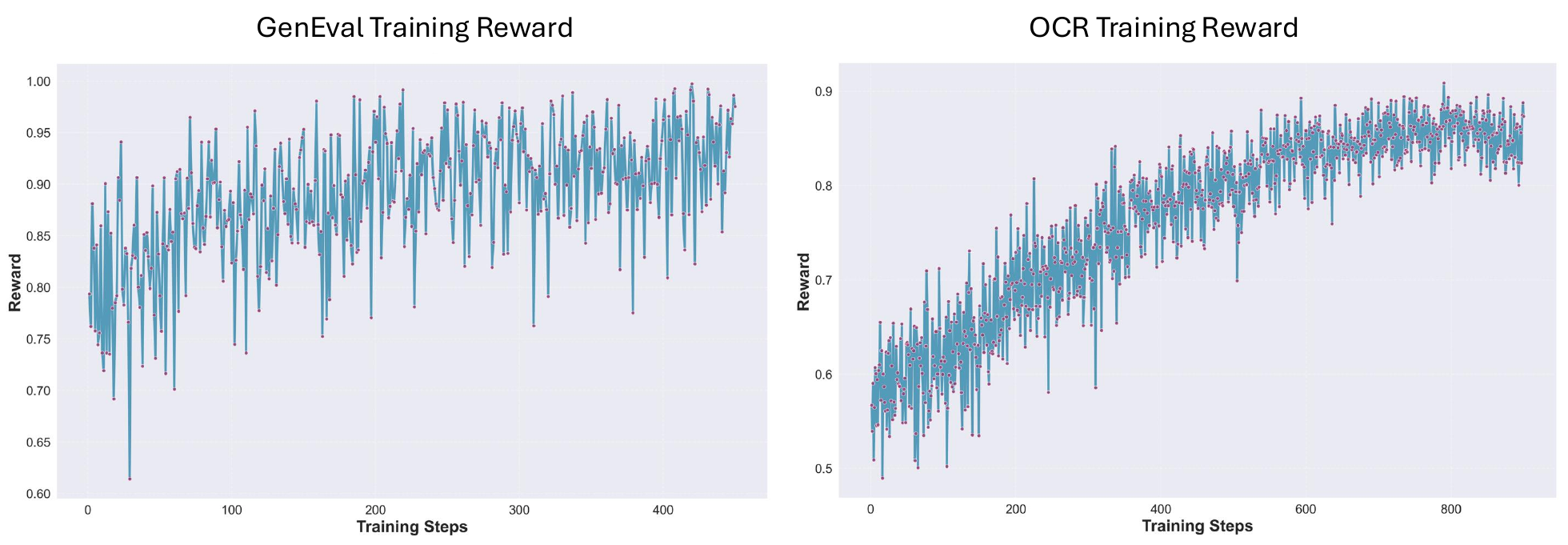}
\vspace{-1em}
\caption{Training reward for GenEval and visual text rendering. 
}
\label{fig:reward}
\vspace{-1em}
\end{figure}

\begin{figure}[!h]
\centering
\includegraphics[width=\linewidth]{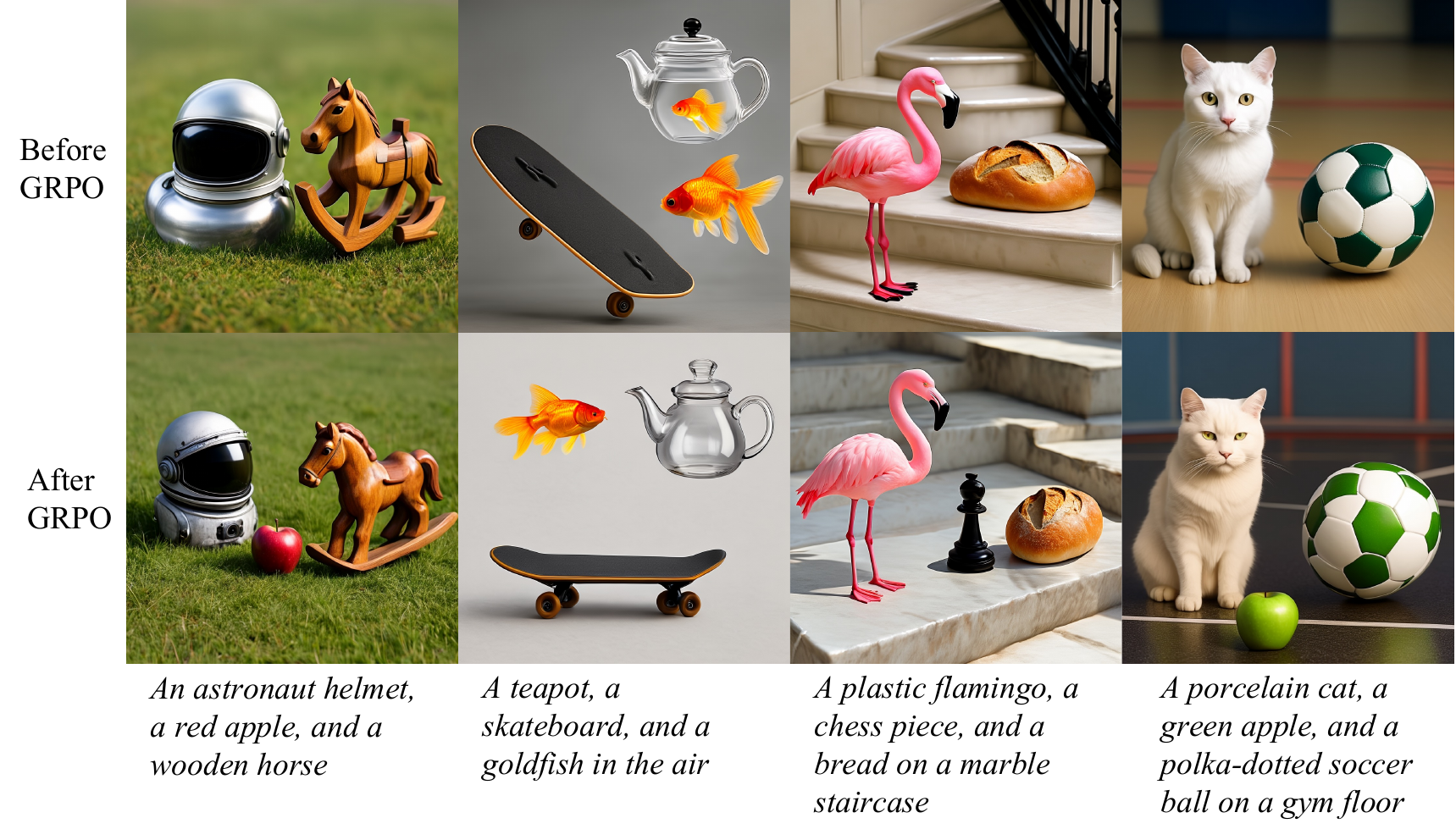}
\caption{ Qualitative results of multiple object composition before and after GRPO.
}
\label{fig:grpo_object}
\end{figure}

\begin{figure}[!h]
\centering
\includegraphics[width=\linewidth]{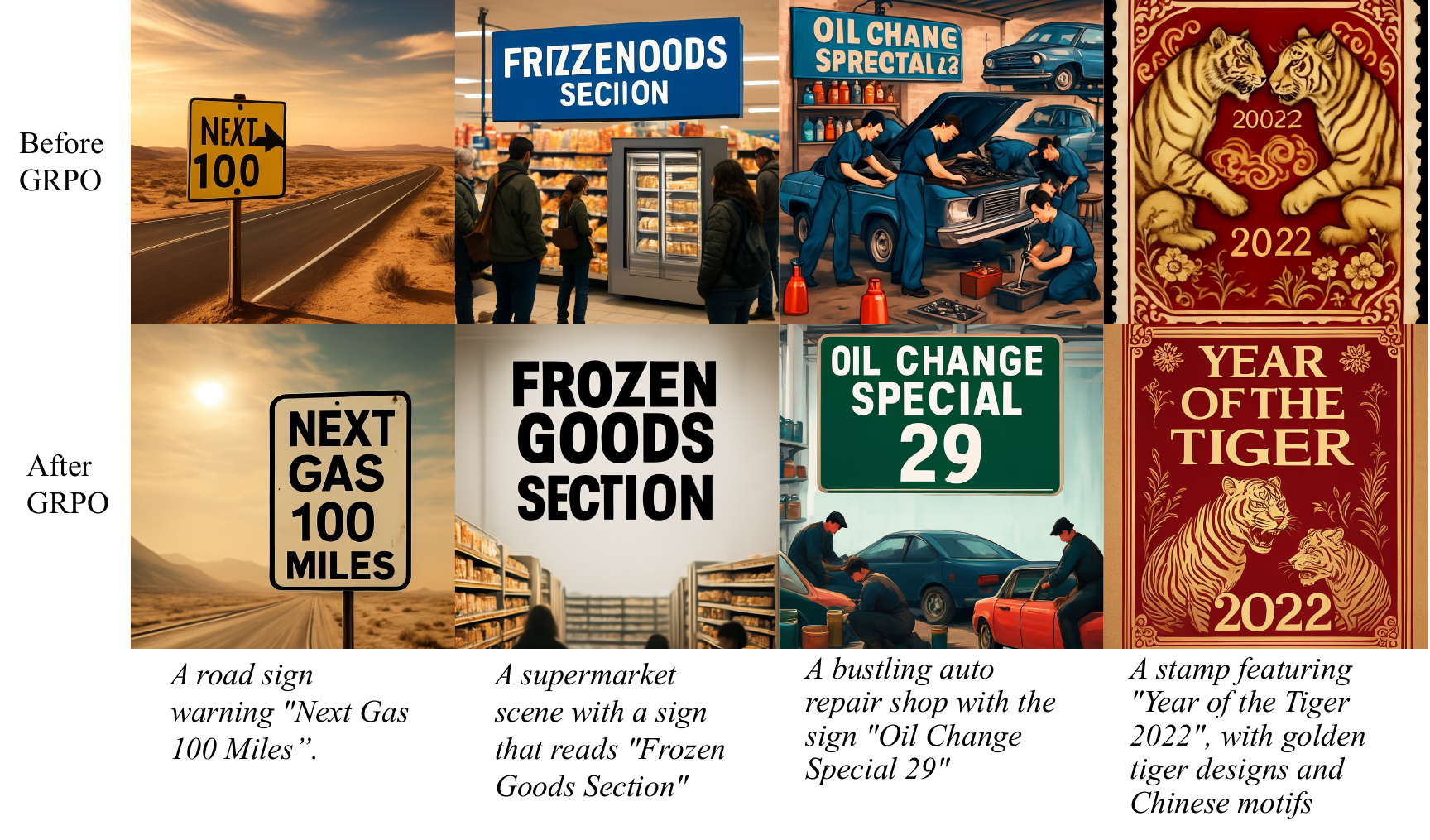}
\vspace{-1em}
\caption{Qualitative results of text rendering before and after GRPO.
}
\label{fig:grpo_text}
\vspace{-1em}
\end{figure}

In our experiments, we only apply RL for autoregressive model and focus on two verifiable reward tasks.
(1) Multiple object composition: training prompts are adopted from Flow-GRPO~\cite{liu2025flow}, and evaluation is performed with the GenEval evaluator.
(2) Visual text rendering: training prompts are also sourced from Flow-GRPO~\cite{liu2025flow}, and evaluation relies on PaddleOCR~\cite{cui2025paddleocr30technicalreport}. Figure~\ref{fig:reward} clearly illustrates the increasing reward trend throughout training. Figures~\ref{fig:grpo_object} and~\ref{fig:grpo_text} present qualitative comparisons before and after GRPO training, demonstrating noticeable improvements in both object composition and text rendering quality.

\subsection{Discussion}

The model architecture determines whether RL is applied to the autoregressive or diffusion model. For example, in Qwen-Image, the autoregressive module mainly serves as a text encoder, while the diffusion model performs most of the image generation. RL is more effective when applied to the diffusion model for Qwen-Image. In contrast, within the \modelname{} framework, the autoregressive model is responsible for producing image tokens, and the diffusion model primarily functions as an image decoder. The autoregressive model plays a central role in image generation, making it the natural target for RL optimization. Empirically, without diffusion acceleration, applying RL to diffusion model is slower due to the lack of KV cache support and the need for multiple time steps.

\begin{findingbox}
 The central challenge in applying reinforcement learning to native image generation lies in metric design rather than the RL algorithm itself. The key open problem is to develop reward models that can effectively capture and balance multiple dimensions, including image quality, instruction following, and human preference alignment.
\end{findingbox}

\section{Image Editing}

To incorporate reference images as multimodal conditions for image generation, a common strategy is to feed them into the autoregressive backbone. Recent models such as MetaQuery~\cite{pan2025transfer}, BLIP3-o~\cite{chen2025blip3}, OmniGen2~\cite{wu2025omnigen2} and Qwen-Image~\cite{wu2025qwen} are built on top of existing vision–language models such as Qwen-VL~\cite{bai2025qwen2}, which natively support image inputs as the conditions.
Within the \modelname{} framework, this is achieved by converting the reference image into quantized image tokens and concatenating them with the input text prompt tokens.  

The primary challenge in image editing is maintaining consistency between the generated and reference images. To address this issue, we adopt the following strategies to enhance consistency.

\paragraph{Image Reconstruction Task.}
We perform an image reconstruction task, where the reference image is provided as input and the text prompt is “\textit{Keep the image unchanged.}”. This task encourages the model to faithfully reconstruct visual details and align the generative process with the conditioning image.

\paragraph{Conditioning on VAE Latents.}

While the reference image is represented through a semantic vision encoder to autoregressive model, such representations often lack fine-grained pixel-level information. To address this limitation, we incorporate low-level, detail-preserving VAE latents as additional conditioning signals for the diffusion model. Specifically, we explore two complementary methods for integrating these VAE latents: (1) cross-attention conditioning and (2) noise-space injection, both designed to enhance visual consistency with respect to the reference image. Empirically, we find that combining both approaches yields the best consistency.

\begin{figure}[!t]
\centering
\includegraphics[width=\linewidth]{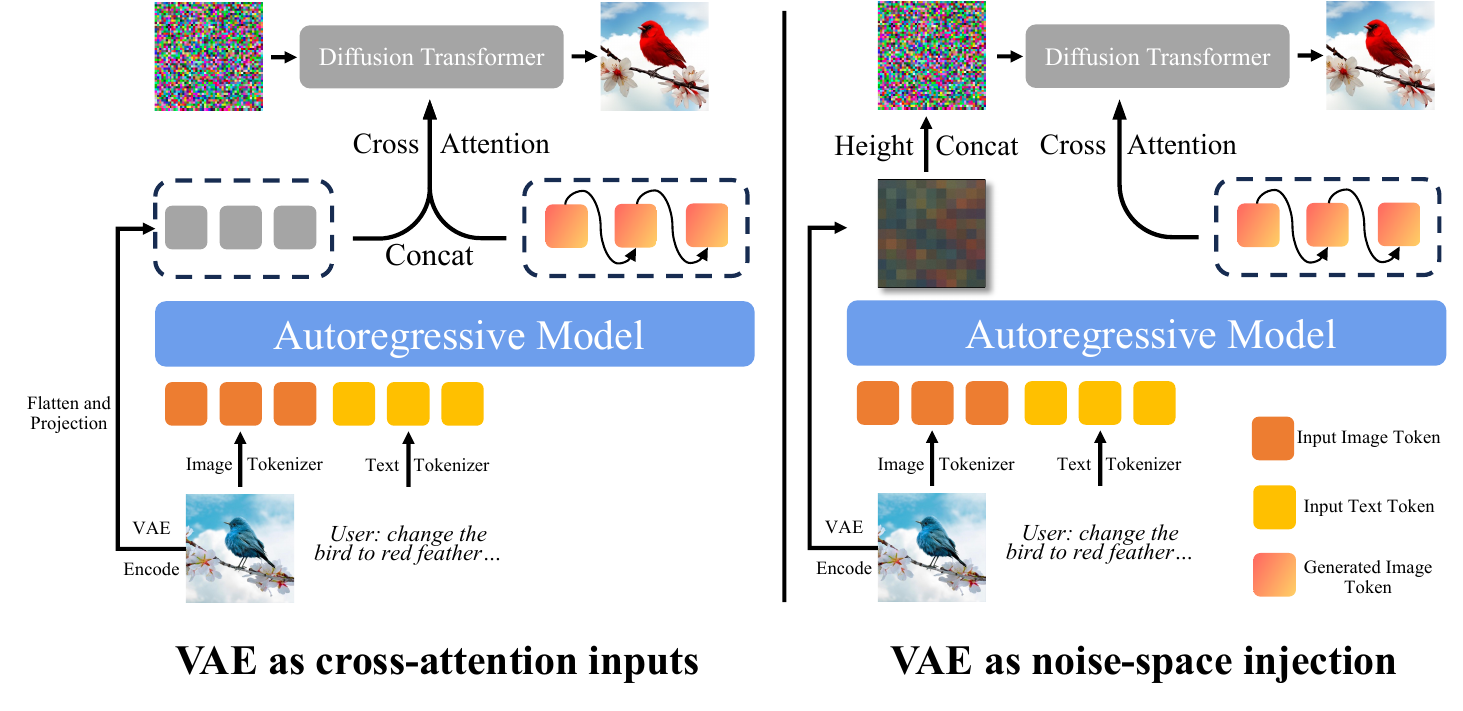}
\caption{Comparison of VAE feature integration strategies in \modelname{}. In the \textit{VAE as cross-attention inputs} setup, the flattened VAE tokens are appended to the multimodal tokens produced by the autoregressive model. Empirically, we find that combining both methods yields the best visual consistency.}
\label{fig:image_edit}
\end{figure}

\textbf{(1) VAE features as cross-attention inputs in DiT.}
After extracting the reference image’s VAE features, we flatten the spatial dimensions (height and width) and project the channel dimension to match the cross-attention input size of the DiT. The resulting projected features are then concatenated with the multimodal context tokens produced by the autoregressive model. These concatenated tokens serve as the input to the DiT’s cross-attention layers, enabling the model to incorporate both semantic and low-level cues, as shown in Figure~\ref{fig:image_edit} left.

\textbf{(2) VAE features as noise-space injection.}
Alternatively, we concatenate the extracted VAE features with the diffusion noise tensor along the height dimension, effectively augmenting the noise input with adjacent reference image VAE features. This composite input is then fed into the DiT during denoising. After the model predicts the refined noise, we compute the loss only over the region corresponding to the original random noise, using it as the flow-matching loss objective as shown in Figure~\ref{fig:image_edit} right.

\begin{table*}[t]
\centering
\small
\setlength{\tabcolsep}{1.1pt}
\begin{tabular}{lccccccccc|c}
\toprule
\textbf{Model} & Add & Adjust & Extract & Replace & Remove & Background & Style & Hybrid & Action & Overall $\uparrow$ \\
\midrule
MagicBrush~\cite{zhang2023magicbrush} & 2.84 & 1.58 & 1.51 & 1.97 & 1.58 & 1.75 & 2.38 & 1.62 & 1.22 & 1.90 \\
Instruct-Pix2Pix~\cite{brooks2023instructpix2pix} & 2.45 & 1.83 & 1.44 & 2.07 & 1.50 & 1.44 & 3.55 & 1.20 & 1.46 & 1.88 \\
AnyEdit~\cite{yu2025anyedit} & 3.18 & 2.95 & 1.88 & 2.47 & 2.23 & 2.24 & 2.85 & 1.56 & 2.65 & 2.45 \\
OmniGen~\cite{xiao2025omnigen} & 3.67 & 3.39 & 1.71 & 2.94 & 2.43 & 3.41 & 4.19 & 2.24 & 3.38 & 2.93 \\
ICEdit~\cite{zhang2025context} & 3.39 & 3.39 & 1.73 & 3.15 & 2.93 & 3.08 & 3.62 & 2.09 & 3.06 & 2.95 \\
StepX-Edit~\cite{liu2025step1x} & 3.88 & 3.14 & 1.76 & 3.40 & 2.41 & 3.16 & 4.63 & 2.64 & 2.52 & 3.06 \\
BAGEL~\cite{deng2025emerging} & 3.76 & 3.04 & 1.70 & 3.43 & 3.04 & 3.40 & 4.64 & 2.64 & 2.62 & 3.25 \\
UniWorld-V1~\cite{lin2025uniworld} & 3.82 & 3.64 & 2.27 & 3.47 & 3.24 & 2.99 & 4.21 & 2.96 & 2.74 & 3.26 \\
OmniGen2~\cite{wu2025omnigen2} & 3.57 & 3.06 & 1.77 & 3.74 & 3.02 & 3.57 & 4.81 & 2.52 & 4.68 & 3.44 \\
FLUX.1 Kontext (Pro)~\cite{labs2025flux} & 4.15 & 2.35 & 4.56 & 3.57 & 3.42 & 4.56 & 4.57 & 3.63 & 4.63 & 4.00 \\
GPT Image 1~\cite{gpt4o} & 4.61 & 4.33 & 2.90 & 4.35 & 3.66 & 4.57 & 4.93 & 3.96 & 4.89 & 4.20 \\
Qwen-Image~\cite{wu2025qwen} & 4.38 & 4.16 & 3.43 & 4.66 & 4.14 & 4.38 & 4.81 & 3.82 & 4.69 & 4.27 \\
\rowcolor{green!10}  \modelname{} (3B) & 4.00 & 3.78 & 2.39 & 4.05 & 2.61 & 4.30 & 4.64 & 2.67 & 4.13 & 3.62 \\
\bottomrule
\end{tabular}
\caption{Image editing benchmark results for ImgEdit~\cite{ye2025imgedit}, with all metrics evaluated using GPT-4.1. The “Overall” score is computed as the average across all task categories. While our 3B model still lags behind GPT-Image and Qwen-Image, it achieves comparable performance to BAGEL and OmniGen2.}
\label{tab:imgedit}
\end{table*}







\subsection{Experiments} 

\begin{figure}[!h]
\centering
\includegraphics[width=0.9\linewidth]{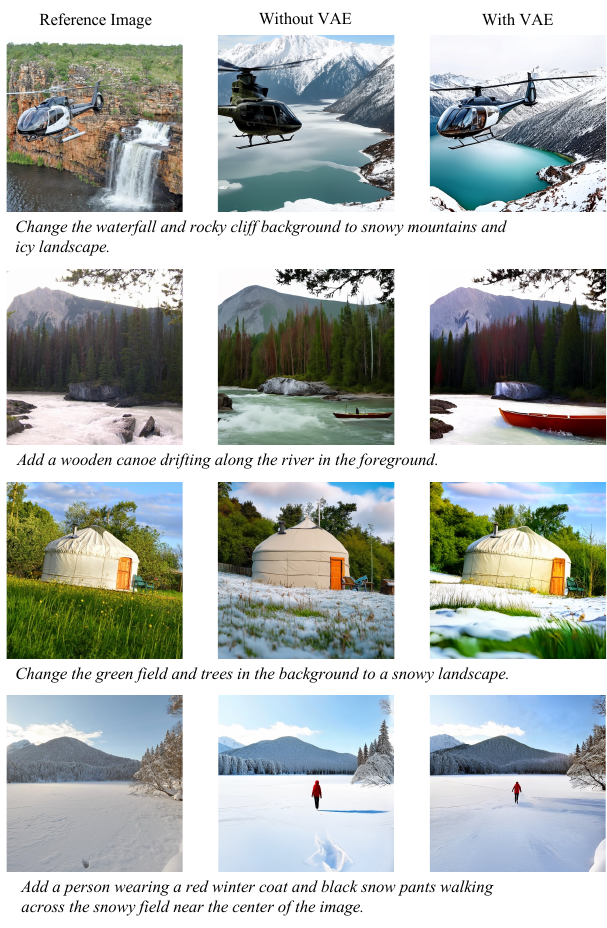}
\caption{Qualitative results for image editing comparing models with and without VAE latent condition.}
\label{fig:editing}
\end{figure}

In our image editing experiments, we adopt a multitask learning setup that jointly trains the model on both image reconstruction and image editing objectives.
The training data is curated from various open-source datasets, including BLIP3-o~\cite{chen2025blip3}, OmniGen2~\cite{wu2025omnigen2}, ShareGPT4V~\cite{chen2023sharegpt4vimprovinglargemultimodal}, and Awesome-Nano-Banana~\cite{awesome-nano-banana}.
To increase the data scale and stabilize training, we repeat selected subsets to construct a larger dataset ensemble.
The final training corpus contains approximately 10 million samples, including repeated entries. Table~\ref{tab:imgedit} presents the results on the ImgEdit~\cite{ye2025imgedit} benchmark, where our 3B model achieves performance comparable to larger models such as BAGEL and OmniGen2. Figure~\ref{fig:editing} illustrates qualitative results on consistency. The comparison between models with and without VAE latents demonstrates that incorporating VAE latents effectively enhances consistency. However, since the VAE in SANA uses a downsampling ratio of 32 to accelerate training and inference~\cite{xie2024sana}, the generated images still exhibit slight inconsistencies with the reference images.

\subsection{Future Exploration}
Besides image reconstruction and conditioning on VAE latents, we also want to highlight the following strategies:

\paragraph{Reinforcement Learning for Image Editing.}
Applying RL to image editing offers a promising direction for improving instruction following and consistency between the generated output and the reference input. While model-based reward functions can effectively guide instruction following~\cite{wu2025editreward}, the design of reward models for measuring consistency remains relatively underexplored.

\paragraph{Designing System Prompts for Inpainting and Subject-Driven Generation.}
Another avenue of exploration is the design of system prompts that explicitly distinguishes between inpainting and subject-driven generation tasks. Inpainting tasks prioritize spatial and background consistency, whereas subject-driven generation emphasizes maintaining consistency in the subject’s appearance. Therefore, incorporating task-specific system prompts can effectively guide the model to handle these two categories of editing more appropriately.

\paragraph{Prompt Rewriting.}
In practice, we use prompt rewriting to enrich prompt details and improve instruction following abilities for both image generation and editing tasks.

\begin{findingbox}
While Autoregressive + Diffusion architectures enable strong instruction following, editing consistency still lags behind, even with VAE feature injected into the diffusion model. Bridging this gap calls for advances in data engineering and scaling reinforcement learning specifically tailored for image editing.
\end{findingbox}

\section{Training Recipe and Evaluation}

Data quality remains a decisive factor in determining the overall performance of the model. Although different models have different data engine pipelines, they generally share several key components. (1) Diversity: We categorize image topics into domains such as Environments, Business, Cities, Food \& Drink, Nature, Objects, Pets, Wildlife, and Lifestyle. The dataset integrates publicly available sources, including CC12M~\cite{changpinyo2021conceptual}, SA-1B~\cite{kirillov2023segment}, and JourneyDB~\cite{JourneyDB}, supplemented with additional proprietary images. (2) Filtering: We perform extensive data cleaning; for example, we remove extremely low-resolution or corrupted images, and we exclude samples containing watermarks, etc. (3) Captioning: Dense captions are generated using Qwen-VL-2.5~\cite{bai2025qwen2}, samples with overly long captions (exceeding 120 tokens) or low CLIP-based image-text alignment scores are discarded. (4) Synthetic data: To enrich training diversity, we further construct synthetic datasets, particularly for text-rendering tasks, and distill data from frontier models.

Regarding the evaluation, although there are numerous benchmarks for evaluating image generation performance~\cite{ghosh2023geneval, hu2024ella, niu2025wise}, there remains a significant need for more specialized benchmarks, particularly for image editing. Such benchmarks are essential for assessing a model’s instruction following ability and the consistency between the generated images and the reference inputs.

\section{Conclusion}
In this work, we introduced \modelname{}, a fully open-source foundation model that advances the frontier of native image generation by unifying text-to-image synthesis and image editing within a single architecture. Through extensive exploration, we identify four central insights: architectural simplicity coupled with scalability is often sufficient; reinforcement learning holds strong potential to further improve generation quality; post-training remains crucial for instruction following and editing consistency; and high-quality, large-scale data continues to set the performance ceiling.
Building on these principles, \modelname{} integrates the instruction following and reasoning capabilities of autoregressive modeling with the fine-grained rendering ability of diffusion, yielding coherent and high-fidelity visual outputs. Experimental results across diverse benchmarks demonstrate that \modelname{} achieves superior performance in both generation and editing tasks, confirming the effectiveness of the Autoregressive + Diffusion paradigm.

Looking forward, we believe the findings presented here point toward promising directions for the next frontier of foundation models, where unified architectures, reinforcement learning, and scalable post-training jointly drive progress in controllable, instruction-aligned, and high-quality native image generation systems.

\appendix


\newpage

\bibliography{main}
\bibliographystyle{plain}
\end{document}